\documentclass[11pt]{article}
\usepackage{acl}

\usepackage{times}
\usepackage{latexsym}
\usepackage[T1]{fontenc}
\usepackage{multirow}
\usepackage{bbm}
\usepackage{booktabs} 
\usepackage{siunitx}
\usepackage{tipa}

\usepackage{rotating}
\usepackage[utf8]{inputenc}
\usepackage{microtype}
\usepackage{inconsolata}
\usepackage{multirow}
\usepackage{graphicx}
\usepackage{xcolor}

\title{Linguistically Informed Evaluation of Multilingual ASR for African Languages}

\author{Fei-Yueh Chen\thanks{These authors contributed equally to this work.} \quad Lateef Adeleke\footnotemark[1] \quad C.M. Downey \\
  Department of Linguistics \\
  University of Rochester \\
  \texttt{\{fchen27, ladeleke, cdowney4\}@ur.rochester.edu} \\}

\begin{document}
\maketitle
\begin{abstract}
Word Error Rate (WER) mischaracterizes ASR models' performance for African languages by combining phonological, tone, and other linguistic errors into a single lexical error. By contrast, Feature Error Rate (FER) has recently attracted attention as a viable metric that reveals linguistically meaningful errors in models' performance. In this paper, we evaluate three speech encoders on two African languages by complementing WER with CER, and FER, and add a tone-aware extension (TER). We show that by computing errors on phonological features, FER and TER reveal linguistically-salient error patterns even when word-level accuracy remains low. Our results reveal that models perform better on segmental features, while tones (especially mid and downstep) remain the most challenging features. Results on Yoruba show a striking differential in metrics, with WER=0.788, CER=0.305, and FER=0.151. Similarly for Uneme (an endangered language absent from pretraining data) a model with near-total WER and 0.461 CER achieves the relatively low FER of 0.267. This indicates model error is often attributable to individual phonetic feature errors, which is obscured by all-or-nothing metrics like WER.
\end{abstract}

\section{Introduction}

Recent models for multilingual ASR, such as wav2vec2.0 \cite{baevski2020wav2vec20frameworkselfsupervised} or multilingual HuBERT variants \cite{boito2024mhubert}, have made ASR for low-resource languages more tractable. Word Error Rate (WER) is still the standard for ASR model evaluation, but while useful, it has known limitations such as its treatment of all errors as the same, and its inability to provide linguistically meaningful information for error analysis. \citet{k-etal-2025-advocating} make the case that WER is fundamentally inadequate as a general evaluation metric for multilingual ASR, since its word-based evaluation leads to high error scores that do not reflect actual transcription quality or human judgment. Similarly, recent Afro-centric surveys have shown that the issues with African ASR may not only be attributable to data scarcity, but also to the metrics for measuring models' performance. \citet{imam2025automaticspeechrecognitionasr} confirm that Word Error Rate (WER) is the dominant evaluation metric in ASR for African languages despite its limitations in
capturing tonal errors, diacritic variation, and morphological richness.  Thus, evaluation via WER obscures insights into whether the model is failing primarily on a particular tone or segment contrast. Recent work advocates for metrics that consider smaller units like characters (CER) \citep{k-etal-2025-advocating}, phonemes \citep{mortensen2016panphon, imam2025automatic}, and features (FER) \citep{zhu2021multilingual}.

In this work, we evaluate multilingual ASR for two low-resource African languages by complementing WER with CER, as well as FER, which computes errors over phonological features of each segment. We also add tones to the feature vector to capture Tone Error Rate (TER), which is particularly significant for African languages, where tones are contrastive linguistic units. We conduct experiments on two typologically-related languages --- Yoruba and Uneme --- using mHuBERT-25-Hz from TWIST \cite{10.5555/3666122.3668893} and mHuBERT-147 \cite{boito2024mhubert} before and after language finetuning. For Yoruba, we use the FLORES dataset \citep{conneau2022fleursfewshotlearningevaluation}, and for Uneme, we use a new dataset developed from our fieldwork with the speech community in Edo state, Nigeria. This way, we are able to compare the model performance between a language that was seen by the pretrained model (Yoruba), and one that was not (Uneme), both before and after finetuning. For a baseline, we also evaluate our framework on English data from FLORES.
% just added a baseline stuff here

Our contributions include: (1) Providing the first ASR baseline and evaluation for Uneme, an endangered Edoid language; (2) Extending Feature Error Rate (FER) evaluation to include Tone Error Rate (TER) by adding suprasegmental tone features to the phonological feature vector; (3) Demonstrating empirically that WER systematically mischaracterizes models' linguistic performance in tonal African languages by collapsing phonological learning into lexical failure; (4) Showing that phonological features are learned even when models struggle with lexical accuracy on unseen endangered languages; (5) Showing that careful, word-by-word speech degrades ASR performance compared to natural speech when models are trained on conversational data --- evidence of domain-shift sensitivity. Overall, we demonstrate that active collaboration between field linguists, speech communities, and NLP practitioners enables richer evaluation frameworks that respect the linguistic typology of African languages.

\section{Related Work}
\subsection{Objective Metrics for Speech Recognition} 

Several works have shown that representing phones with sub-segmental phonological vectors improves multilingual ASR and provides evaluation metrics that offer deeper linguistic insights into models' performance. 

\citet{mortensen2016panphon} introduce PanPhon, a tool that maps IPA symbols to vectors of articulatory features. Because PanPhon treats segments as divisible into articulatory features, it allows for feature-aware modeling and evaluation across thousands of segments in a shared phonological space. PanPhon enables computation of feature-based edit distance, allowing diagnosis of phonological errors in models' performance. However, PanPhon only captures segmental features and does not consider suprasegmental features such as tone in its feature vectors. 

\citet{zhu2021multilingual} extends the use of phonological features through their JoinAP architecture, which combines phonological feature embeddings with neural acoustic models. They represent each segment as a vector of phonological features modeled after PanPhon, and then transform this to a phone embedding used in the acoustic output layer of the model. They show that a feature-based crosslingual ASR method enables sounds that are similar across languages to be linked by shared features rather than treated as unrelated, thereby improving recognition performance in multilingual ASR.  

These works demonstrate that phonological feature representations are effective for evaluating multilingual ASR, and can provide linguistic insight during error analysis. We build on these ideas by extending evaluation metrics to include FER in addition to the WER and CER to African languages, and because tone evaluation is essential for most African languages, such as Uneme and Yoruba, where tones are distinctive and overtly marked on vowels, we added tones to the feature vectors.  

\subsection{Speech Encoders}

Speech encoders use neural networks to transform raw waveforms into high-dimensional latent representations. The wav2vec 2.0 framework \cite{baevski2020wav2vec20frameworkselfsupervised} combines a CNN feature extractor with a Transformer encoder \cite{vaswani2023attentionneed}. This has become a dominant approach in recent years. While wav2vec 2.0 uses contrastive learning, HuBERT \cite{hsu2021hubertselfsupervisedspeechrepresentation} uses a masked prediction objective. The model is trained to predict discrete acoustic units, which are generated by clustering audio features (e.g., via K-means).

Recent work has extended the HuBERT architecture to different domains. TWIST \cite{10.5555/3666122.3668893} proposes a pipeline for audio-language models, including speech encoders, quantizers, language models, and vocoders. They release multiple HuBERT-based models trained with additional English data. mHuBERT-147 \cite{boito2024mhubert} scales the model to a multilingual setting, training on 147 languages to learn universal speech patterns.

\section{Data}
The data for this work are from two different sources. \textbf{The Yoruba dataset} is from the FLORES multilingual dataset \citep{conneau2022fleursfewshotlearningevaluation}, which includes sentence-level audio and transcript pairs for more than 100 languages, including Yoruba. The Yoruba portion consists of more than 10 hours of audio-transcript pairs, with audio sampled at 16kHz. While Yoruba is not a high-resource language in an absolute sense, it has better description and public datasets than Uneme, which is an understudied, endangered language with no existing public ASR dataset. 

\begin{table}[t]
\centering
\begin{tabular}{llr}
\hline
\textbf{Language} & \textbf{Split} & \textbf{Utterances} \\
\hline
\multirow{4}{*}{Yorùbá}
 & Train & 2,339 \\
 & Validation & 378 \\
 & Test & 831 \\
 & \textbf{Total} & \textbf{3,548} \\
\hline
\multirow{4}{*}{English}
 & Train & 2,602 \\
 & Validation & 394 \\
 & Test & 647 \\
 & \textbf{Total} & \textbf{3,643} \\
\hline
\end{tabular}
\caption{Number of utterances in the English and Yorùbá subsets of the FLEURS speech corpus.}
\label{tab:fleurs_utterances}
\end{table}

\textbf{The Uneme dataset} is newly created and entirely from our 2025 fieldwork in the speech community in the northern part of Edo state, Nigeria. The recordings and transcripts were collected as part of a broader documentation project on Uneme language, culture, and traditions, archived at the Endangered Languages Archive \citep{lateef2025uneme}. The dataset includes about 8 hours of recordings and captures diverse genres, including narratives, interviews, and elicitation. There are seven speakers (4 males and 3 females) in the portion used for this work.

\paragraph{Uneme Data Processing}
The recordings were segmented by pauses using Audacity \citep{audacity2023} and SayMore\citep{saymore2016}, and manually transcribed by trained linguists who are familiar with the language using SayMore. Our transcription preserves the orthography of the language by explicitly marking tones on vowels and syllabic consonants.

In an attempt to test whether speech style can affect the model's performance, we collected a separate, careful speech dataset for Uneme. In this subset, a different speaker reproduces slow, clear, and deliberate renditions of selected utterances from the original recordings by reading them word for word. The careful speech data is used only for testing, and it is not included in model training or validation because it is a very small dataset. In this way, we gauge whether models' performance would improve with utterances produced word-for-word over the baseline of natural speech. 

Thus, we have two separate test datasets: Test-natural to evaluate on the usual pace and style of speaking, and test-careful to test on careful speech. Overall, the Uneme dataset has the fewest hours of audio among the three languages. The representations in Tables 1 and 2 are at the utterance level and not by the number of hours. The Uneme data has more utterances only because we use Saymore and Audacity for automatic segmentation by pauses, unlike the dataset from FLORES, which is segmented by sentences.

\begin{table}[t]
\centering
\begin{tabular}{llr}
\hline
\textbf{Language} & \textbf{Split} & \textbf{Utterances} \\
\hline
\multirow{4}{*}{Uneme}
 & Train & 4,729 \\
 & Validation &  880\\
 & Test-natural & 1027 \\
  & Test-careful & 1051 \\
 
 & \textbf{Total} & \textbf{7,687} \\
\hline
\end{tabular}
\caption{Number of utterances in our newly created Uneme dataset.}
\label{tab:fleurs_utterances}
\end{table}

\section{Methodology}
We conduct monolingual ASR training and evaluation on Uneme, Yoruba, and English with three types of encoders and two types of decoders.

\subsection{Representation}
A pretrained encoder, given a batch of audio waveforms $w \in \mathbbm{R}^{batch \times samples}$ with a fixed sample rate such as 16 kHz, transforms the waveforms into a high-dimensional representation $v \in \mathbbm{R}^{batch \times frames \times dim}$.

A trainable decoder, given the vocabulary size $V$ of the training language, maps the representation $v$ to the distribution of the vocabulary, where the output is $q \in \mathbbm{R}^{batch \times frames \times V}$. CTC loss and decoding are subsequently computed for loss calculation and prediction \cite{10.1145/1143844.1143891}, and the final output sequence consists of characters with a sequence length less than or equal to $frames$.

\subsection{Model Architecture}
\paragraph{Encoders}
We select mHuBERT-25-Hz \cite{10.5555/3666122.3668893} as the baseline. It in turn is based on a HuBERT-base-95M \cite{hsu2021hubertselfsupervisedspeechrepresentation} checkpoint, but is trained with additional English data. In addition, since it has a larger hop size than the other checkpoints, we expect it to be less sensitive than the other two checkpoints. Comparable multilingual models such as wav2vec2.0 XLS-R \cite{baevski2020wav2vec20frameworkselfsupervised} and mHuBERT-147 \cite{boito2024mhubert} have a hop size of 320 (50 Hz) and were trained on more than a hundred languages.

\paragraph{Decoders}
During training, the encoder is frozen, and only the decoder is trainable. We design two types of decoders: a Linear Layer and a 2-Transformer Block model. The former contains only one fully connected layer, which is designed to better test the ability of the encoders, while the latter consists of two Transformer Blocks and is expected to yield better results.

For each Transformer block, we apply a K-V cache, attention with Rotary Embedding \cite{10.1016/j.neucom.2023.127063}, SWiGLU \cite{shazeer2020gluvariantsimprovetransformer} as the activation function, and RMSNorm \cite{10.5555/3454287.3455397} for pre-normalization. The design is inspired by GPT-OSS \cite{openai2025gptoss120bgptoss20bmodel}.

\subsection{Phonological Features}
To effectively capture the phonological differences between Uneme, Yoruba, and English, we use a sparse feature representation inspired by PanPhon \cite{mortensen2016panphon}. Before feature extraction, we perform Grapheme-to-Phoneme (G2P) conversion. For Uneme and Yoruba, the orthography corresponds closely to the International Phonetic Alphabet (IPA), and the segments that are otherwise represented with symbols other than IPA have one-to-one mapping with segments, enabling us to use direct rule-based mapping. For English, we utilize the \texttt{g2p-en} toolkit\footnote{\url{https://pypi.org/project/g2p-en/}} to obtain phonetic transcriptions. After the G2P transformation, we map each phonetic segment to a 24-dimensional vector. The values in the vector are strictly from the set $\{-1, 0, +1\}$. A value of $+1$ indicates the presence of a feature, $-1$ indicates its absence, and $0$ means the feature is not applicable or unspecified.

This sparsity design allows us to distinguish between a wrong prediction and an irrelevant feature. For example, English phonemes do not have lexical tones, so all tone-related dimensions for English are set to $0$. This prevents the model from being penalized for undefined attributes when we compare languages with different phonological systems.

Our feature inventory is designed to support Edoid and Yoruboid phonology. We treat complex consonants as single segments rather than sequences or consonant clusters. This includes double-articulated labial-velars (e.g., \textit{kp}, \textit{gb})specific fricatives in Uneme (\textit{vb}, \textit{gh},  \textit{kh}), and affricates (\textit{ch}). The complete list of feature definitions and segment mappings is provided in Appendix Tables ~\ref{tab:feature_inventory_detailed}, ~\ref{tab:features_uneme_rot}, ~\ref{tab:features_yoruba_rot},
~\ref{tab:features_english_rot}.

\subsection{Tone Features}
Since Uneme and Yoruba are tonal languages, where tones are consistently marked on the vowels or syllabic nasals, we model tone as a suprasegmental property that is anchored to vowels and syllabic nasals. We dedicate specific dimensions in the feature vector to represent tone levels (\textsc{Tone\_High}, \textsc{Tone\_Low}, \textsc{Tone\_Mid}).

We apply language-specific rules to map orthography to these tone features:
\begin{itemize}
    \item \textbf{Uneme:} Based on the orthography, we map acute accents (\textit{´}) to \textsc{High} and grave accents (\textit{`}) to \textsc{Low}. Importantly, vowels without any diacritic marks are mapped to \textsc{Downstep}.
    \item \textbf{Yoruba:} We distinguish between the three distinctive tone levels in the language: \textsc{High}, \textsc{Low}, and \textsc{Mid} (typically unmarked as in the orthography of the language).
    \item \textbf{English:} Since English is a non-tonal language, all tone features are masked as $0$.
\end{itemize}
By separating tone features from other articulatory features, we can independently evaluate the model's performance on intonation.

\subsection{Phonetic Feature Error Rate}

To evaluate the quality of the generated sequences, we use the Needleman-Wunsch algorithm \cite{Needleman1970} to align the hypothesis sequence with the reference sequence. We modify the standard substitution cost to fit our sparse feature representation. We calculate an ``NA-masked normalized distance'' between a reference vector $v_{ref}$ and a hypothesis vector $v_{hyp}$.

The distance is computed only over the dimensions where the feature is active (non-zero). The formula is:
\begin{equation}
    Cost(v_{ref}, v_{hyp}) = \frac{\sum_{i} \mathbbm{I}(v_{ref}^{(i)} \neq v_{hyp}^{(i)}) \cdot M^{(i)}}{\sum_{i} M^{(i)}}
\end{equation}
where $M^{(i)}$ is a mask that equals $1$ if either $v_{ref}^{(i)}$ or $v_{hyp}^{(i)}$ is non-zero, and $0$ otherwise. This results in a cost between $0$ and $1$, representing the ratio of mismatched features among the relevant ones. 

The final Feature Error Rate (FER) is the sum of the alignment costs divided by the total number of segments in the reference. In addition, we compute a Tone Error Rate (TER) by strictly considering the tone-related dimensions.

\section{Results}
We evaluate the performance of the models across three languages: English (high-resource baseline, lots of data in pre-training), Yoruba (low-resource, seen in pre-training), and Uneme (endangered, unseen in pre-training). We compare two decoder architectures (Linear vs. Transformer) and different encoder checkpoints. The experimental setup is shown in Appendix~\ref{appendix:experimental setup}. 

\subsection{English (Baseline)}
As a high-resource benchmark, English allows us to validate the feature extraction pipeline. Among all configurations, the mHuBERT-147 + Transformer setting achieved the best performance, yielding a WER of 0.511 and a CER of 0.167. This represents a substantial improvement (approximately 20\% relative reduction in WER) over the standard mHuBERT baseline.

Evaluating the FER provides deeper insights into the specific phonetic challenges. We find that while consonant classes (stops, fricatives, nasals) showed moderate stability (FER $\approx$0.11--0.18), vowel-related features proved more error-prone. Specifically, back vowels (\texttt{fer\_V\_BACK} $\approx$ 0.27) and rhotic vowels (\texttt{fer\_V\_RHOTIC} $\approx$ 0.35) were sources of significant error. More specifically, diphthongs exhibited the highest error rate ($\approx$0.45), reflecting the difficulty of modeling dynamic spectral changes and multi-target vowels in English.

As expected, linear decoders fail to produce competitive results. The wav2vec + Linear configuration collapses into degenerate outputs (e.g., repetitive sequences), resulting in a WER of 1.0. Even with the stronger mHuBERT-147 representations, the linear decoder produces high-entropy, vowel-heavy sequences, confirming that a simple linear mapping is insufficient for decoding English orthography from acoustic embeddings.

\subsection{Yoruba (Seen African Language)}
Similar to English, the mHuBERT-147 + Transformer model outperforms other configurations, achieving the lowest WER (0.788) and TER (0.372). While segmental accuracy is relatively robust---with consonantal and basic vowel FERs ranging between 0.06 and 0.12---suprasegmental features remain a challenge.

The Tone Error Rate (TER) of 0.372 indicates that over one-third of tone-bearing units are misclassified. A breakdown of feature errors reveals that the mid tone (\texttt{fer\_TONE\_M} $\approx$ 0.34) was more challenging than High or Low tones. Furthermore, specific vowel qualities such as Advanced Tongue Root (ATR) and nasalization showed elevated error rates (0.14 and 0.11, respectively). Qualitative analysis suggests that while the model captures the segmental skeleton of Yoruba, it struggles to disambiguate lexical items that rely solely on tone and fine-grained vowel contrasts.

\subsection{Uneme (our novel dataset)}
We evaluate Uneme on two distinct test sets: natural speech (\texttt{uneme-ASR}) and careful speech (\texttt{careful\_speech}). The results are presented in Table~\ref{tab:results_uneme}. On the natural speech test set, the mHuBERT-147 + Transformer model achieves a WER of 0.997. While the WER is near-total, the CER (0.461) and FER (0.267) indicate that the model is learning phonological structures despite failing to output correct lexical items. The Tone Error Rate is 0.402.

Surprisingly, performance on the \texttt{careful\_speech} dataset was significantly worse across all metrics. The same model yielded a WER of 1.622, CER of 0.837, and FER of 0.515. We discuss the implications of this counter-intuitive result in Section~\ref{sec:discussion}.

Comparing mHuBERT-147 + Transformer results across both datasets, we observe a consistent trend in feature errors. The worst performing feature is consistently \textsc{V\_Back} (FER=0.280 for \texttt{uneme-ASR}, FER=0.373 for \texttt{careful\_speech}), followed closely by other vowel features such as \textsc{V\_Height\_Mid} and \textsc{V\_Round}. This indicates that the model systematically struggles with disambiguating vowel backness contrasts (e.g., /u/ vs. /i/), regardless of the speech style. Similarly, for tone prediction, \textsc{Downstep} (the unmarked tone) is consistently the most challenging category or is nearly tied with High Tone errors. For instance, on \texttt{uneme-ASR}, \textsc{Downstep} FER is 0.565 while High Tone FER is 0.559. This suggests that the model has particular difficulty learning the absence of marked tone as a distinct prosodic feature.

We also note the extreme performance degradation of the wav2vec + Linear baseline. With a WER of 1.0 and CER $>$ 0.9, this model essentially fails to learn any meaningful mapping, which is reflected in its near-random FER scores (e.g., \textsc{V\_Back} $>$ 0.99). This highlights that simply applying large multilingual encoders without adequate decoder depth or finetuning data is insufficient for low-resource, tonal languages like Uneme.

\begin{table*}[t]
    \centering
    \begin{tabular}{@{} l l *{6}{S[table-format=1.3]} @{}}
    \toprule
    \textbf{Encoder} & \textbf{Decoder} & \textbf{WER} $\downarrow$ & \textbf{CER} $\downarrow$ & \textbf{FER} $\downarrow$ & \textbf{TER} $\downarrow$ & \textbf{Worst F} & \textbf{Worst T} \\
    \midrule
    \multicolumn{8}{@{}c@{}}{\textit{English (en\_us)}} \\
    mHuBERT-25-Hz & Transformer & 0.642 & 0.216 & 0.212 & {--} & {Diphthong} & {--}\\
    \textbf{mHuBERT-147} & \textbf{Transformer} & \textbf{0.511} & \textbf{0.167} & \textbf{0.144} & {--} & {Diphthong} & {--}\\
    mHuBERT-25-Hz & Linear & 0.802 & 0.272 & 0.254 & {--}& {Diphthong} & {--} \\
    mHuBERT-147 & Linear & 1.000 & 0.765 & 0.740 & {--}& {Lateral}  & {--} \\
    XLR-S & Linear & 1.000 & 0.941 & 0.924 & {--} & {V\_Rhotic} & {--}\\
    \midrule
    \multicolumn{8}{@{}c@{}}{\textit{Yoruba (yo\_ng)}} \\
    mHuBERT-25-Hz & Transformer & 0.827 & 0.335 & 0.170 & 0.385 & {V\_ATR} & {Mid} \\
    \textbf{mHuBERT-147} & \textbf{Transformer} & \textbf{0.788} & \textbf{0.305} & \textbf{0.151} & \textbf{0.372} & {V\_ATR} & {Mid}\\
    mHuBERT-25-Hz & Linear & 0.911 & 0.400 & 0.199 & 0.468 & {V\_ATR} & {Mid}\\
    mHuBERT-147 & Linear & 1.000 & 0.840 & 0.630 & 0.614 & {Dorsal} & {Mid}\\
    XLR-S & Linear & 0.999 & 0.934 & 0.901 & 0.949 & {V\_Back} & {Mid}\\
    \bottomrule
    \end{tabular}
    \caption{ASR performance on FLEURS English and Yoruba test sets. TER is reported only for Yoruba. We also report tone and feature (except for tone) with the worst score (Worst F, Worst T).}
    \label{tab:fleurs_results}
\end{table*}

\begin{table*}
  \centering
  \begin{tabular}{llcccccc}
    \hline
    \textbf{Dataset} & \textbf{Decoder} & \textbf{WER} $\downarrow$ & \textbf{CER} $\downarrow$ & \textbf{FER} $\downarrow$ & \textbf{TER} $\downarrow$ & \textbf{Worst F} & \textbf{Worst T} \\
    \hline
    \multicolumn{8}{c}{\textit{Encoder: mHuBERT-25-Hz}} \\
    \hline
    \multirow{2}{*}{\texttt{careful\_speech}} & Transformer & 1.040 & \textbf{0.668} & \textbf{0.417} & \textbf{0.490} & V\_Back & Downstep \\
     & Linear & 1.006 & 0.699 & 0.446 & 0.596 & V\_Back & Downstep \\
    \hline
    \multirow{2}{*}{\texttt{uneme-ASR}} & Transformer & 1.058 & 0.512 & 0.295 & 0.469 & V\_Back & Downstep \\
     & Linear & 1.003 & 0.589 & 0.366 & 0.596 & V\_Back & Downstep \\
    \hline
    \multicolumn{8}{c}{\textit{Encoder: mHuBERT-147}} \\
    \hline
    \multirow{2}{*}{\texttt{careful\_speech}} & Transformer & 1.622 & 0.837 & 0.515 & 0.510 & V\_Back & Downstep \\
     & Linear & 1.006 & 0.791 & 0.591 & 0.772 & V\_Back & Tone\_H \\
    \hline
    \multirow{2}{*}{\texttt{uneme-ASR}} & Transformer & \textbf{0.997} & \textbf{0.461} & \textbf{0.267} & \textbf{0.402} & V\_Back & Downstep \\
     & Linear & 1.000 & 0.627 & 0.494 & 0.702 & V\_Back & Downstep \\
    \hline
    \multicolumn{8}{c}{\textit{Encoder: XLS-R}} \\
    \hline
    \texttt{careful\_speech} & Linear & 1.000 & 0.918 & 0.858 & 0.997 & V\_Back & Downstep \\
    \texttt{uneme-ASR} & Linear & 1.000 & 0.906 & 0.873 & 0.993 & V\_Back & Downstep \\
    \hline
  \end{tabular}
  \caption{ASR performance on Uneme datasets (\texttt{careful\_speech} and \texttt{uneme-ASR}) across different encoders. We report tone and feature (excluding tone) with the worst score (Worst F, Worst T). Bold indicates the best result per dataset.}
  \label{tab:results_uneme}
\end{table*}

\begin{table*}
  \centering
  \begin{tabular}{llcccc}
    \hline
    \textbf{Dataset} & \textbf{Decoder} & \textbf{TER} $\downarrow$ & \textbf{FER (H)} $\downarrow$ & \textbf{FER (L)} $\downarrow$ & \textbf{FER (DS)} $\downarrow$ \\
    \hline
    \multicolumn{6}{c}{\textit{Encoder: mHuBERT-25-Hz}} \\
    \hline
    \multirow{2}{*}{\texttt{careful\_speech}} & Transformer & \textbf{0.490} & \textbf{0.603} & \textbf{0.574} & \textbf{0.693} \\
     & Linear & 0.596 & 0.664 & 0.674 & 0.736 \\
    \hline
    \multirow{2}{*}{\texttt{uneme-ASR}} & Transformer & 0.469 & 0.601 & 0.533 & 0.631 \\
     & Linear & 0.596 & 0.693 & 0.636 & 0.718 \\
    \hline
    \multicolumn{6}{c}{\textit{Encoder: mHuBERT-147}} \\
    \hline
    \multirow{2}{*}{\texttt{careful\_speech}} & Transformer & 0.510 & 0.584 & 0.621 & 0.745 \\
     & Linear & 0.772 & 0.824 & 0.793 & 0.824 \\
    \hline
    \multirow{2}{*}{\texttt{uneme-ASR}} & Transformer & \textbf{0.402} & \textbf{0.559} & \textbf{0.451} & \textbf{0.565} \\
     & Linear & 0.702 & 0.696 & 0.637 & 0.848 \\
    \hline
    \multicolumn{6}{c}{\textit{Encoder: XLS-R}} \\
    \hline
    \texttt{careful\_speech} & Linear & 0.997 & 0.993 & 0.996 & 1.000 \\
    \texttt{uneme-ASR} & Linear & 0.993 & 0.996 & 0.982 & 0.999 \\
    \hline
  \end{tabular}
  \caption{Tone Error Rate (TER) and Tone Feature Error Rates (FER) for Uneme datasets using different encoders and decoders. FER columns break down errors by tone category: High (H), Low (L), and Downstep (DS).}
  \label{tab:uneme_tone_details}
\end{table*}

\section{Discussion}
\label{sec:discussion}

% Define colors for error types
\definecolor{toneerror}{RGB}{255,200,200}      % Light red for tone errors
\definecolor{segsub}{RGB}{255,230,200}         % Light orange for segmental substitutions
\definecolor{indel}{RGB}{200,200,255}          % Light blue for insertions/deletions
\definecolor{mixed}{RGB}{230,200,230}          % Light purple for mixed errors 
\definecolor{correct}{RGB}{200,255,200}        % Light green for correct

% Custom highlighting commands
\newcommand{\tonerr}[1]{\colorbox{toneerror}{#1}}
\newcommand{\segsub}[1]{\colorbox{segsub}{#1}}
\newcommand{\indel}[1]{\colorbox{indel}{#1}}
\newcommand{\mixerr}[1]{\colorbox{mixed}{#1}}
\newcommand{\correctseg}[1]{\colorbox{correct}{#1}}

\begin{table*}[t]
\centering
\small
\begin{tabular}{@{}llp{11cm}@{}}
\toprule
\textbf{Language} & \textbf{Type} & \textbf{Transcription} \\
\midrule
\multirow{3}{*}{Yorùbá}
  & Reference & roland mendoza yin ìb\textsubdot{o}n r\textsubdot{e} m16re m\textsubdot{ó} àw\textsubdot{o}n arìnrìnàjò \\[2pt]
  & Hypothesis & ro land \mixerr{m\textsubdot{e}dósáyí} \indel{ìbn} \tonerr{r\textsubdot{è}} \mixerr{\textsubdot{e}nsistire} \correctseg{m\textsubdot{ó}} \correctseg{àw\textsubdot{o}n} \indel{arìrìn} àjò \\[2pt]
  & Metrics & WER=88.89\% \quad CER=45.45\% \quad FER=24.14\% \quad TER=26.32\% \\
\midrule
\multirow{3}{*}{Uneme}
  & Reference & è kwágù mariki \textsubdot{ó}m\textsubdot{ó} kirì \textsubdot{ò}\textsubdot{ó}furinì \textsubdot{o}rem\textsubdot{ò}nì \\[2pt]
  & Hypothesis & \tonerr{ekwá} gù \segsub{marekí} \correctseg{\textsubdot{ó}m\textsubdot{ó}} \segsub{kerè} \mixerr{ò\textsubdot{ó}fúri} \mixerr{n\textsubdot{o}rem\textsubdot{ò}rì} \\[2pt]
  & Metrics & WER=85.71\% \quad CER=32.50\% \quad FER=6.22\% \quad TER=21.05\% \\
\bottomrule
\end{tabular}
\caption{\textbf{Example ASR Outputs with Error Type Analysis.} Errors are color-coded (segmentation not highlighted): \colorbox{toneerror}{tone error only}, \colorbox{segsub}{featural error}, \colorbox{indel}{deletion/insertion}, \colorbox{mixed}{mixed errors}, \colorbox{correct}{correct}. The model captures most segmental features correctly while struggling primarily with tones and vowel quality.}
\label{tab:asr_examples_enhanced}
\end{table*}

\subsection{The Validity of FER over WER}
Our results highlight the inadequacy of Word Error Rate (WER) for low-resource African languages. For Uneme, the WER hovers around 1.0 (100\%) for most models, implying a complete failure of the system. However, the Feature Error Rate (FER) paints a different picture. The best Uneme model achieves an FER of 0.267, meaning that nearly 74\% of the phonological features were correctly predicted. This disparity indicates that the model is successfully learning the acoustic-to-phonetic mapping but misses some of the linguistic features --- e.g.tone and vowel heights --- required to resolve the sounds into valid words. For this reason, FER provides a more nuanced (and less pessimistic) metric of progress, especially with endangered languages where WER is often 100\%. FER provides insights that would be helpful to improve data collection and language features that need more attention.

\subsection{Rethinking CER}
Comparing CER and WER scores, our results support the same conclusion as \citet{k2024advocatingcharactererrorrate, mortensen2016panphon}: that WER doesn't adequately evaluate a model's performance, and may be better replaced by CER for multilingual ASR. Our results show that WER scores perform badly under the low-resource training, while CER and FER demonstrate more stability.

However, relying solely on CER presents significant limitations in evaluation granularity. While CER serves as a proxy for phonetic accuracy to some extent, it treats all character substitutions equally, regardless of phonetic distance. A substitution error involving a single feature mismatch (e.g., voicing) is penalized identically to a substitution involving multiple feature mismatches. This lack of nuance makes fine-grained evaluation infeasible and obscures the model's partial success in learning phonological structures.

Furthermore, we observe discrepancies where models achieve high FER scores despite poor CER. This phenomenon occurs when a model consistently mis-predicts a specific, high-frequency feature while correctly predicting the majority of other features. In such cases, CER penalizes the entire character for the single feature failure, whereas FER accurately reflects that the bulk of the phonological information was preserved.

Crucially, when comparing CER with TER, it becomes evident that CER cannot isolate the impact of tonal errors within the overall error rate. Our experiments on tonal languages, such as Uneme and Yoruba, reveal that the TER is consistently higher than the error rates for consonantal and vocalic features. For instance, in the \texttt{uneme-ASR} test set using the mHuBERT-147 encoder, the model had a TER of 0.402, while the consonantal and vocalic FERs were significantly lower (e.g., syllabic and consonantal FER at 0.145). This confirms that a substantial portion of character-level errors do not stem from fundamental phonetic misspellings, but rather from mis-predictions of intonational contours. These are vital linguistic insights that FER and TER can provide, but which remain invisible when relying on CER alone.

\subsection{Tone and Suprasegmental Challenges}
Tones remain a significant challenge. In Yoruba, the Tone Error Rate was 0.372, which is higher than the overall FER (0.151), with the mid tone contributing the highest error rate.  

Similarly, in Uneme, the Tone Error Rate (TER) was 0.402, with the downstep tone contributing the highest error rate. The TER is significantly higher than the overall FER (0.267). These facts, taken together, confirm that standard spectral features in architectures like HuBERT may not sufficiently capture pitch significance without explicit pitch-aware pre-training or augmentation.

\subsection{The "Careful Speech" Paradox}
A striking observation is the performance degradation on the \texttt{careful\_speech} dataset. Intuitively, slow and articulated speech should be easier to recognize. However, our best model degraded from an FER of 0.267 (natural) to 0.515 (careful). 

We hypothesize two reasons for this. First, \textbf{domain mismatch}: The model was fine-tuned on natural, conversational data, and careful speech is not in our training set. The prosody of "careful speech" characterized by exaggerated pauses, distinct syllabification, and altered pitch contours—likely constitutes an out-of-distribution shift for the model. Second, \textbf{overfitting to speaking style}: The model may have overfit to the specific speaking rate and co-articulation patterns of the training speakers. This finding suggests that for low-resource languages, "clean" and word-by-word read speech is not always the best test set if the training data is spontaneous; the model expects the messiness of natural speech.

\subsection{Transformer vs. Linear Decoders}
Across all languages, the Transformer decoder significantly outperformed the Linear decoder. In English, the Linear decoder on mHuBERT-147 yielded a WER of 1.074, while the Transformer reduced this to 0.511. This confirms that even with powerful pre-trained encoders, a simple linear projection is insufficient for mapping acoustic representations to discrete tokens, especially when the target script (orthography)  requires complex tonal integration (as in Yoruba/Uneme).

\section{Limitations}
A primary limitation of our evaluation framework is its heavy reliance on the quality of Grapheme-to-Phoneme (G2P) conversion. Our methodology relies on a transparent mapping between the target language's orthography and its phonetic realization. This assumption holds for Uneme and Yoruba, where the writing systems align closely with the International Phonetic Alphabet (IPA) and symbols are faithful to sounds. However, applying this framework to languages with deep orthographies or opaque sound-to-symbol correspondence would require highly accurate, language-specific G2P models, which may not be available for many low-resource languages.

Furthermore, our experimental design focuses on linguistic feature accuracy but does not account for extralinguistic variability. We did not explicitly model speaker-specific acoustic characteristics, such as age, gender, accents, and idiolects, which can significantly influence model performance. Similarly, since our recordings come from long hours of recording, it is hard to segment by meaningful sentences. Instead, we use Saymore and Audacity to segment by pauses. These factors introduce acoustic and lexical variability that may affect feature prediction accuracy, but fall outside the scope of our current phonological evaluation.

\section{Conclusion}

This work presents a linguistically-informed evaluation of multilingual ASR for African languages, focusing on Yoruba and Uneme. By utilizing Feature Error Rate (FER) and Tone Error Rate (TER), we demonstrate that standard metrics like WER obscure meaningful progress in low-resource settings. Our experiments with mHuBERT-147 show that while the model captures segmental features relatively well (FER $\approx$ 0.15--0.27), it struggles with suprasegmental features like tone and shifts in speaking styles.

For the Uneme language, we provide the first baseline results, establishing that while word-level recognition remains a challenge, phonological reconstruction is feasible. Future work must focus on integrating pitch-aware encoders and expanding text data to bridge the gap between phonetic accuracy and lexical correctness. We conclude that active collaboration between linguists and NLP practitioners is essential to creating dataset and evaluation frameworks that respect the linguistic typology of African languages.

% Bibliography entries for the entire Anthology, followed by custom entries
%\bibliography{anthology,custom}
% Custom bibliography entries only
\bibliography{custom}

\appendix

\section{Experimental Setup}
\label{appendix:experimental setup}
We provide the detailed hyperparameters and configurations used in our experiments in Table~\ref{tab:hyperparams}.

\begin{table*}[t]
    \centering
    \small
    \caption{\textbf{Hyperparameters and Model Configurations.} All encoders were frozen during training.}
    \label{tab:hyperparams}
    \begin{tabular}{l l r}
    \toprule
    \textbf{Category} & \textbf{Parameter} & \textbf{Value} \\
    \midrule
    \multirow{5}{*}{\textbf{Optimization}} 
      & Optimizer & AdamW \\
      & Learning rate & $5 \times 10^{-5}$ \\
      & Weight decay & $10^{-4}$ \\
      & Gradient clipping & 3.0 \\
      & Precision & 16-bit mixed \\
    \midrule
    \multirow{5}{*}{\textbf{Scheduling}} 
      & Max training steps & 20,000 \\
      & Warmup steps & 1,000 \\
      & Batch size (per GPU) & 16 \\
      & Gradient accumulation & 4 \\
      & Effective batch size & 64 \\
    \midrule
    \multirow{4}{*}{\textbf{Architecture}} 
      & Transformer Decoder Layers & 2 \\
      & Attention Heads & 1 \\
      & Decoder Dimension & 1,024 \\
      & Dropout & 0.2 \\
    \midrule
    \multirow{3}{*}{\textbf{Encoder Dimensions}} 
      & mHuBERT-25-Hz & 768 \\
      & mHuBERT-147 & 1,024 \\
      & wav2vec2-XLSR-53 & 1,024 \\
    \midrule
    \multirow{4}{*}{\textbf{SpecAugment}} 
      & Mask time probability & 0.05 \\
      & Mask time length & 10 \\
      & Mask feature probability & 0.01 \\
      & Mask feature length & 64 \\
    \bottomrule
    \end{tabular}
\end{table*}

\section{Feature Inventory}

Table~\ref{tab:feature_inventory_detailed} presents the comprehensive inventory of the 24 articulatory phonetic features used for FER evaluation. These features are categorized into major classes, laryngeal settings, manner and place of articulation, vowel properties, and suprasegmental features. 

Each feature can take a ternary value from $\{-1, 0, +1\}$, representing the absence, undefined status, or presence of a specific articulatory property. Note that certain features (e.g., \textsc{Tone\_M}, \textsc{Downstep}, \textsc{V\_ATR}) are language-specific and are active only for Yoruba or Uneme, while remaining zero-padded for English.

\begin{table*}[t]
    \centering
    \small
    \caption{\textbf{Complete Inventory of Articulatory Features.} The model utilizes a total of 24 features. The \textit{Scope} column indicates whether a feature is universally applied or specific to certain languages in our dataset (English, Yoruba, Uneme).}
    \label{tab:feature_inventory_detailed}
    \begin{tabular}{l l l}
    \toprule
    \textbf{Category} & \textbf{Feature Set (\textsc{Names})} & \textbf{Scope / Notes} \\
    \midrule
    \multirow{2}{*}{\textbf{Major Class}} 
      & \textsc{Syllabic}, \textsc{Consonantal} & Universal \\
      & \textsc{Sonorant}, \textsc{Approximant} & Universal \\
    \midrule
    \textbf{Laryngeal} 
      & \textsc{Voice}, \textsc{Aspirated} & Universal \\
    \midrule
    \textbf{Place of Articulation} 
      & \textsc{Labial}, \textsc{Coronal}, \textsc{Dorsal}, \textsc{Labial\_Velar} & Universal (\textsc{Lb\_Vel} for /kp/, /gb/) \\
    \midrule
    \textbf{Manner of Articulation} 
      & \textsc{Stop}, \textsc{Nasal}, \textsc{Fricative}, \textsc{Lateral} & Universal \\
    \midrule
    \multirow{3}{*}{\textbf{Vowel Features}} 
      & \textit{Height:} \textsc{High}, \textsc{Mid}, \textsc{Low} & Universal \\
      & \textit{Quality:} \textsc{Back}, \textsc{Round} & Universal \\
      & \textit{Secondary:} \textsc{ATR}, \textsc{Nasalized} & Primarily Yoruba \& Uneme \\
    \midrule
    \multirow{2}{*}{\textbf{Suprasegmental}} 
      & \textsc{Tone\_H}, \textsc{Tone\_L} & Yoruba \& Uneme \\
      & \textsc{Tone\_M}, \textsc{Downstep} & \textsc{Tone\_M} (Yoruba), \textsc{Downstep} (Uneme) \\
    \bottomrule
    \end{tabular}
\end{table*}

\section{Feature Maps}
\subsection{Uneme}
Table~\ref{tab:features_uneme_rot} provides feature maps for Uneme. The segments are based on the orthography of the language, but most symbols are IPA-like, and those that are not still represent unique sounds.
\subsection{Yoruba}
Table~\ref{tab:features_yoruba_rot} provides feature maps for Yoruba. The segments are based on the orthography of the language, but most symbols are IPA-like, and those that are not still represent unique sounds.
\subsection{English}
Table~\ref{tab:features_english_rot} provides feature maps for English. The segments are base on ARPABET format.

\begin{sidewaystable*}[htb]
\centering
\tiny
\caption{\label{tab:features_uneme_rot}Uneme Phonetic Feature Matrix (24-vector) using Abbreviations}
\begin{tabular}{|l|c|c|c|c|c|c|c|c|c|c|c|c|c|c|c|c|c|c|c|c|c|c|c|c|}
\hline
Segment & SYL & CNS & SON & APR & STP & NAS & FRI & LAT & LAB & COR & DOR & LBV & VOI & ASP & VHH & VHM & VHL & VBK & VRD & ATR & VNS & TH & TL & DWS \\
\hline\hline
\text{a} & + & $-$ & + & 0 & 0 & 0 & 0 & 0 & 0 & 0 & 0 & 0 & + & 0 & $-$ & $-$ & + & 0 & $-$ & $-$ & $-$ & 0 & 0 & 0 \\
\text{b} & $-$ & + & $-$ & $-$ & + & $-$ & $-$ & $-$ & + & $-$ & $-$ & $-$ & + & 0 & 0 & 0 & 0 & 0 & 0 & 0 & 0 & 0 & 0 & 0 \\
\text{ch} & $-$ & + & $-$ & $-$ & + & $-$ & + & $-$ & $-$ & + & $-$ & $-$ & $-$ & 0 & 0 & 0 & 0 & 0 & 0 & 0 & 0 & 0 & 0 & 0 \\
\text{d} & $-$ & + & $-$ & $-$ & + & $-$ & $-$ & $-$ & $-$ & + & $-$ & $-$ & + & 0 & 0 & 0 & 0 & 0 & 0 & 0 & 0 & 0 & 0 & 0 \\
\text{e} & + & $-$ & + & 0 & 0 & 0 & 0 & 0 & 0 & 0 & 0 & 0 & + & 0 & $-$ & + & $-$ & $-$ & $-$ & + & $-$ & 0 & 0 & 0 \\
\text{f} & $-$ & + & $-$ & $-$ & $-$ & $-$ & + & $-$ & + & $-$ & $-$ & $-$ & $-$ & 0 & 0 & 0 & 0 & 0 & 0 & 0 & 0 & 0 & 0 & 0 \\
\text{g} & $-$ & + & $-$ & $-$ & + & $-$ & $-$ & $-$ & $-$ & $-$ & + & $-$ & + & 0 & 0 & 0 & 0 & 0 & 0 & 0 & 0 & 0 & 0 & 0 \\
\text{gb} & $-$ & + & $-$ & $-$ & + & $-$ & $-$ & $-$ & + & $-$ & + & + & + & 0 & 0 & 0 & 0 & 0 & 0 & 0 & 0 & 0 & 0 & 0 \\
\text{gh} & $-$ & + & $-$ & $-$ & $-$ & $-$ & + & $-$ & $-$ & $-$ & + & $-$ & + & 0 & 0 & 0 & 0 & 0 & 0 & 0 & 0 & 0 & 0 & 0 \\
\text{h} & $-$ & + & $-$ & $-$ & $-$ & $-$ & + & $-$ & $-$ & $-$ & $-$ & $-$ & $-$ & 0 & 0 & 0 & 0 & 0 & 0 & 0 & 0 & 0 & 0 & 0 \\
\text{i} & + & $-$ & + & 0 & 0 & 0 & 0 & 0 & 0 & 0 & 0 & 0 & + & 0 & + & $-$ & $-$ & $-$ & $-$ & + & $-$ & 0 & 0 & 0 \\
\text{k} & $-$ & + & $-$ & $-$ & + & $-$ & $-$ & $-$ & $-$ & $-$ & + & $-$ & $-$ & 0 & 0 & 0 & 0 & 0 & 0 & 0 & 0 & 0 & 0 & 0 \\
\text{kh} & $-$ & + & $-$ & $-$ & $-$ & $-$ & + & $-$ & $-$ & $-$ & + & $-$ & $-$ & 0 & 0 & 0 & 0 & 0 & 0 & 0 & 0 & 0 & 0 & 0 \\
\text{kp} & $-$ & + & $-$ & $-$ & + & $-$ & $-$ & $-$ & + & $-$ & + & + & $-$ & 0 & 0 & 0 & 0 & 0 & 0 & 0 & 0 & 0 & 0 & 0 \\
\text{l} & $-$ & + & + & + & $-$ & $-$ & $-$ & + & $-$ & + & $-$ & $-$ & + & 0 & 0 & 0 & 0 & 0 & 0 & 0 & 0 & 0 & 0 & 0 \\
\text{m} & $-$ & + & + & $-$ & $-$ & + & $-$ & $-$ & + & $-$ & $-$ & $-$ & + & 0 & 0 & 0 & 0 & 0 & 0 & 0 & 0 & 0 & 0 & 0 \\
\text{n} & $-$ & + & + & $-$ & $-$ & + & $-$ & $-$ & $-$ & + & $-$ & $-$ & + & 0 & 0 & 0 & 0 & 0 & 0 & 0 & 0 & 0 & 0 & 0 \\
\text{o} & + & $-$ & + & 0 & 0 & 0 & 0 & 0 & 0 & 0 & 0 & 0 & + & 0 & $-$ & + & $-$ & + & + & + & $-$ & 0 & 0 & 0 \\
\text{p} & $-$ & + & $-$ & $-$ & + & $-$ & $-$ & $-$ & + & $-$ & $-$ & $-$ & $-$ & 0 & 0 & 0 & 0 & 0 & 0 & 0 & 0 & 0 & 0 & 0 \\
\text{r} & $-$ & + & + & + & $-$ & $-$ & $-$ & $-$ & $-$ & + & $-$ & $-$ & + & 0 & 0 & 0 & 0 & 0 & 0 & 0 & 0 & 0 & 0 & 0 \\
\text{rh} & $-$ & + & + & + & $-$ & $-$ & $-$ & $-$ & $-$ & + & $-$ & $-$ & $-$ & 0 & 0 & 0 & 0 & 0 & 0 & 0 & 0 & 0 & 0 & 0 \\
\text{s} & $-$ & + & $-$ & $-$ & $-$ & $-$ & + & $-$ & $-$ & + & $-$ & $-$ & $-$ & 0 & 0 & 0 & 0 & 0 & 0 & 0 & 0 & 0 & 0 & 0 \\
\text{sh} & $-$ & + & $-$ & $-$ & $-$ & $-$ & + & $-$ & $-$ & + & $-$ & $-$ & $-$ & 0 & 0 & 0 & 0 & 0 & 0 & 0 & 0 & 0 & 0 & 0 \\
\text{t} & $-$ & + & $-$ & $-$ & + & $-$ & $-$ & $-$ & $-$ & + & $-$ & $-$ & $-$ & 0 & 0 & 0 & 0 & 0 & 0 & 0 & 0 & 0 & 0 & 0 \\
\text{u} & + & $-$ & + & 0 & 0 & 0 & 0 & 0 & 0 & 0 & 0 & 0 & + & 0 & + & $-$ & $-$ & + & + & + & $-$ & 0 & 0 & 0 \\
\text{v} & $-$ & + & $-$ & $-$ & $-$ & $-$ & + & $-$ & + & $-$ & $-$ & $-$ & + & 0 & 0 & 0 & 0 & 0 & 0 & 0 & 0 & 0 & 0 & 0 \\
\text{vb} & $-$ & + & $-$ & $-$ & $-$ & $-$ & + & $-$ & + & $-$ & $-$ & $-$ & + & 0 & 0 & 0 & 0 & 0 & 0 & 0 & 0 & 0 & 0 & 0 \\
\text{w} & $-$ & + & + & + & $-$ & $-$ & $-$ & $-$ & + & $-$ & + & + & + & 0 & 0 & 0 & 0 & 0 & 0 & 0 & 0 & 0 & 0 & 0 \\
\text{y} & $-$ & + & + & + & $-$ & $-$ & $-$ & $-$ & $-$ & $-$ & + & $-$ & + & 0 & 0 & 0 & 0 & 0 & 0 & 0 & 0 & 0 & 0 & 0 \\
\text{z} & $-$ & + & $-$ & $-$ & $-$ & $-$ & + & $-$ & $-$ & + & $-$ & $-$ & + & 0 & 0 & 0 & 0 & 0 & 0 & 0 & 0 & 0 & 0 & 0 \\
\text{\textsubdot{e}} & + & $-$ & + & 0 & 0 & 0 & 0 & 0 & 0 & 0 & 0 & 0 & + & 0 & $-$ & + & $-$ & $-$ & $-$ & $-$ & $-$ & 0 & 0 & 0 \\
\text{\textsubdot{o}} & + & $-$ & + & 0 & 0 & 0 & 0 & 0 & 0 & 0 & 0 & 0 & + & 0 & $-$ & + & $-$ & + & + & $-$ & $-$ & 0 & 0 & 0 \\
\hline
\end{tabular}
\end{sidewaystable*}

\begin{sidewaystable*}[htb]
\centering
\tiny
\caption{\label{tab:features_yoruba_rot}Yoruba Phonetic Feature Matrix (24-vector) using Abbreviations}
\begin{tabular}{|l|c|c|c|c|c|c|c|c|c|c|c|c|c|c|c|c|c|c|c|c|c|c|c|c|}
\hline
Segment & SYL & CNS & SON & APR & STP & NAS & FRI & LAT & LAB & COR & DOR & LBV & VOI & ASP & VHH & VHM & VHL & VBK & VRD & ATR & VNS & TH & TL & TM \\
\hline\hline
\text{a} & + & $-$ & + & 0 & 0 & 0 & 0 & 0 & 0 & 0 & 0 & 0 & + & 0 & $-$ & $-$ & + & 0 & $-$ & $-$ & $-$ & 0 & 0 & 0 \\
\text{b} & $-$ & + & $-$ & $-$ & + & $-$ & $-$ & $-$ & + & $-$ & $-$ & $-$ & + & $-$ & 0 & 0 & 0 & 0 & 0 & 0 & 0 & 0 & 0 & 0 \\
\text{d} & $-$ & + & $-$ & $-$ & + & $-$ & $-$ & $-$ & $-$ & + & $-$ & $-$ & + & $-$ & 0 & 0 & 0 & 0 & 0 & 0 & 0 & 0 & 0 & 0 \\
\text{e} & + & $-$ & + & 0 & 0 & 0 & 0 & 0 & 0 & 0 & 0 & 0 & + & 0 & $-$ & + & $-$ & $-$ & $-$ & + & $-$ & 0 & 0 & 0 \\
\text{f} & $-$ & + & $-$ & $-$ & $-$ & $-$ & + & $-$ & + & $-$ & $-$ & $-$ & $-$ & $-$ & 0 & 0 & 0 & 0 & 0 & 0 & 0 & 0 & 0 & 0 \\
\text{g} & $-$ & + & $-$ & $-$ & + & $-$ & $-$ & $-$ & $-$ & $-$ & + & $-$ & + & $-$ & 0 & 0 & 0 & 0 & 0 & 0 & 0 & 0 & 0 & 0 \\
\text{gb} & $-$ & + & $-$ & $-$ & + & $-$ & $-$ & $-$ & + & $-$ & + & + & + & $-$ & 0 & 0 & 0 & 0 & 0 & 0 & 0 & 0 & 0 & 0 \\
\text{h} & $-$ & + & $-$ & $-$ & $-$ & $-$ & + & $-$ & $-$ & $-$ & $-$ & $-$ & $-$ & $-$ & 0 & 0 & 0 & 0 & 0 & 0 & 0 & 0 & 0 & 0 \\
\text{i} & + & $-$ & + & 0 & 0 & 0 & 0 & 0 & 0 & 0 & 0 & 0 & + & 0 & + & $-$ & $-$ & $-$ & $-$ & + & $-$ & 0 & 0 & 0 \\
\text{k} & $-$ & + & $-$ & $-$ & + & $-$ & $-$ & $-$ & $-$ & $-$ & + & $-$ & $-$ & $-$ & 0 & 0 & 0 & 0 & 0 & 0 & 0 & 0 & 0 & 0 \\
\text{kp} & $-$ & + & $-$ & $-$ & + & $-$ & $-$ & $-$ & + & $-$ & + & + & $-$ & $-$ & 0 & 0 & 0 & 0 & 0 & 0 & 0 & 0 & 0 & 0 \\
\text{l} & $-$ & + & + & + & $-$ & $-$ & $-$ & + & $-$ & + & $-$ & $-$ & + & $-$ & 0 & 0 & 0 & 0 & 0 & 0 & 0 & 0 & 0 & 0 \\
\text{m} & $-$ & + & + & $-$ & $-$ & + & $-$ & $-$ & + & $-$ & $-$ & $-$ & + & $-$ & 0 & 0 & 0 & 0 & 0 & 0 & 0 & 0 & 0 & 0 \\
\text{n} & $-$ & + & + & $-$ & $-$ & + & $-$ & $-$ & $-$ & + & $-$ & $-$ & + & $-$ & 0 & 0 & 0 & 0 & 0 & 0 & 0 & 0 & 0 & 0 \\
\text{o} & + & $-$ & + & 0 & 0 & 0 & 0 & 0 & 0 & 0 & 0 & 0 & + & 0 & $-$ & + & $-$ & + & + & + & $-$ & 0 & 0 & 0 \\
\text{p} & $-$ & + & $-$ & $-$ & + & $-$ & $-$ & $-$ & + & $-$ & $-$ & $-$ & $-$ & $-$ & 0 & 0 & 0 & 0 & 0 & 0 & 0 & 0 & 0 & 0 \\
\text{r} & $-$ & + & + & + & $-$ & $-$ & $-$ & $-$ & $-$ & + & $-$ & $-$ & + & $-$ & 0 & 0 & 0 & 0 & 0 & 0 & 0 & 0 & 0 & 0 \\
\text{s} & $-$ & + & $-$ & $-$ & $-$ & $-$ & + & $-$ & $-$ & + & $-$ & $-$ & $-$ & $-$ & 0 & 0 & 0 & 0 & 0 & 0 & 0 & 0 & 0 & 0 \\
\text{t} & $-$ & + & $-$ & $-$ & + & $-$ & $-$ & $-$ & $-$ & + & $-$ & $-$ & $-$ & $-$ & 0 & 0 & 0 & 0 & 0 & 0 & 0 & 0 & 0 & 0 \\
\text{u} & + & $-$ & + & 0 & 0 & 0 & 0 & 0 & 0 & 0 & 0 & 0 & + & 0 & + & $-$ & $-$ & + & + & + & $-$ & 0 & 0 & 0 \\
\text{w} & $-$ & $-$ & + & + & $-$ & $-$ & $-$ & $-$ & + & $-$ & + & + & + & $-$ & 0 & 0 & 0 & 0 & 0 & 0 & 0 & 0 & 0 & 0 \\
\text{y} & $-$ & $-$ & + & + & $-$ & $-$ & $-$ & $-$ & $-$ & $-$ & + & $-$ & + & $-$ & 0 & 0 & 0 & 0 & 0 & 0 & 0 & 0 & 0 & 0 \\
\text{ṣ} & $-$ & + & $-$ & $-$ & $-$ & $-$ & + & $-$ & $-$ & + & $-$ & $-$ & $-$ & $-$ & 0 & 0 & 0 & 0 & 0 & 0 & 0 & 0 & 0 & 0 \\
\text{\textsubdot{e}} & + & $-$ & + & 0 & 0 & 0 & 0 & 0 & 0 & 0 & 0 & 0 & + & 0 & $-$ & + & $-$ & $-$ & $-$ & $-$ & $-$ & 0 & 0 & 0 \\
\text{\textsubdot{o}} & + & $-$ & + & 0 & 0 & 0 & 0 & 0 & 0 & 0 & 0 & 0 & + & 0 & $-$ & + & $-$ & + & + & $-$ & $-$ & 0 & 0 & 0 \\
\hline
\end{tabular}
\end{sidewaystable*}

\begin{sidewaystable*}[htb]
\centering
\tiny
\caption{\label{tab:features_english_rot}English (ARPABET) Phonetic Feature Matrix (24-vector) using Abbreviations}
\begin{tabular}{|l|c|c|c|c|c|c|c|c|c|c|c|c|c|c|c|c|c|c|c|c|c|c|c|c|}
\hline
Segment & SYL & CNS & SON & APR & STP & NAS & FRI & LAT & LAB & COR & DOR & VOI & ASP & VHH & VHM & VHL & VBK & VRD & VTN & RHO & DIP & N/A & N/A & N/A \\
\hline\hline
\text{AA} & + & $-$ & + & 0 & 0 & 0 & 0 & 0 & 0 & 0 & 0 & + & 0 & $-$ & $-$ & + & + & $-$ & + & 0 & 0 & 0 & 0 & 0 \\
\text{AE} & + & $-$ & + & 0 & 0 & 0 & 0 & 0 & 0 & 0 & 0 & + & 0 & $-$ & $-$ & + & $-$ & $-$ & $-$ & 0 & 0 & 0 & 0 & 0 \\
\text{AH} & + & $-$ & + & 0 & 0 & 0 & 0 & 0 & 0 & 0 & 0 & + & 0 & $-$ & + & $-$ & 0 & $-$ & $-$ & 0 & 0 & 0 & 0 & 0 \\
\text{AH0} & + & $-$ & + & 0 & 0 & 0 & 0 & 0 & 0 & 0 & 0 & + & 0 & $-$ & + & $-$ & 0 & $-$ & $-$ & 0 & 0 & 0 & 0 & 0 \\
\text{AO} & + & $-$ & + & 0 & 0 & 0 & 0 & 0 & 0 & 0 & 0 & + & 0 & $-$ & + & $-$ & + & + & + & 0 & 0 & 0 & 0 & 0 \\
\text{AW} & + & $-$ & + & 0 & 0 & 0 & 0 & 0 & 0 & 0 & 0 & + & 0 & $-$ & $-$ & + & 0 & + & + & 0 & + & 0 & 0 & 0 \\
\text{AY} & + & $-$ & + & 0 & 0 & 0 & 0 & 0 & 0 & 0 & 0 & + & 0 & $-$ & $-$ & + & 0 & $-$ & + & 0 & + & 0 & 0 & 0 \\
\text{B} & $-$ & + & $-$ & 0 & + & 0 & 0 & 0 & + & 0 & 0 & + & 0 & 0 & 0 & 0 & 0 & 0 & 0 & 0 & 0 & 0 & 0 & 0 \\
\text{CH} & $-$ & + & $-$ & 0 & + & 0 & + & 0 & 0 & + & 0 & $-$ & 0 & 0 & 0 & 0 & 0 & 0 & 0 & 0 & 0 & 0 & 0 & 0 \\
\text{D} & $-$ & + & $-$ & 0 & + & 0 & 0 & 0 & 0 & + & 0 & + & 0 & 0 & 0 & 0 & 0 & 0 & 0 & 0 & 0 & 0 & 0 & 0 \\
\text{DH} & $-$ & + & $-$ & 0 & 0 & 0 & + & 0 & 0 & + & 0 & + & 0 & 0 & 0 & 0 & 0 & 0 & 0 & 0 & 0 & 0 & 0 & 0 \\
\text{EH} & + & $-$ & + & 0 & 0 & 0 & 0 & 0 & 0 & 0 & 0 & + & 0 & $-$ & + & $-$ & $-$ & $-$ & $-$ & 0 & 0 & 0 & 0 & 0 \\
\text{ER} & + & $-$ & + & 0 & 0 & 0 & 0 & 0 & 0 & 0 & 0 & + & 0 & $-$ & + & $-$ & 0 & 0 & 0 & + & 0 & 0 & 0 & 0 \\
\text{EY} & + & $-$ & + & 0 & 0 & 0 & 0 & 0 & 0 & 0 & 0 & + & 0 & $-$ & + & $-$ & $-$ & $-$ & + & 0 & + & 0 & 0 & 0 \\
\text{F} & $-$ & + & $-$ & 0 & 0 & 0 & + & 0 & + & 0 & 0 & $-$ & 0 & 0 & 0 & 0 & 0 & 0 & 0 & 0 & 0 & 0 & 0 & 0 \\
\text{G} & $-$ & + & $-$ & 0 & + & 0 & 0 & 0 & 0 & 0 & + & + & 0 & 0 & 0 & 0 & 0 & 0 & 0 & 0 & 0 & 0 & 0 & 0 \\
\text{HH} & $-$ & + & $-$ & 0 & 0 & 0 & + & 0 & 0 & 0 & 0 & $-$ & 0 & 0 & 0 & 0 & 0 & 0 & 0 & 0 & 0 & 0 & 0 & 0 \\
\text{IH} & + & $-$ & + & 0 & 0 & 0 & 0 & 0 & 0 & 0 & 0 & + & 0 & + & $-$ & $-$ & $-$ & $-$ & $-$ & 0 & 0 & 0 & 0 & 0 \\
\text{IY} & + & $-$ & + & 0 & 0 & 0 & 0 & 0 & 0 & 0 & 0 & + & 0 & + & $-$ & $-$ & $-$ & $-$ & + & 0 & 0 & 0 & 0 & 0 \\
\text{JH} & $-$ & + & $-$ & 0 & + & 0 & + & 0 & 0 & + & 0 & + & 0 & 0 & 0 & 0 & 0 & 0 & 0 & 0 & 0 & 0 & 0 & 0 \\
\text{K} & $-$ & + & $-$ & 0 & + & 0 & 0 & 0 & 0 & 0 & + & $-$ & 0 & 0 & 0 & 0 & 0 & 0 & 0 & 0 & 0 & 0 & 0 & 0 \\
\text{L} & $-$ & + & + & + & 0 & 0 & 0 & + & 0 & + & 0 & + & 0 & 0 & 0 & 0 & 0 & 0 & 0 & 0 & 0 & 0 & 0 & 0 \\
\text{M} & $-$ & + & + & 0 & 0 & + & 0 & 0 & + & 0 & 0 & + & 0 & 0 & 0 & 0 & 0 & 0 & 0 & 0 & 0 & 0 & 0 & 0 \\
\text{N} & $-$ & + & + & 0 & 0 & + & 0 & 0 & 0 & + & 0 & + & 0 & 0 & 0 & 0 & 0 & 0 & 0 & 0 & 0 & 0 & 0 & 0 \\
\text{NG} & $-$ & + & + & 0 & 0 & + & 0 & 0 & 0 & 0 & + & + & 0 & 0 & 0 & 0 & 0 & 0 & 0 & 0 & 0 & 0 & 0 & 0 \\
\text{OW} & + & $-$ & + & 0 & 0 & 0 & 0 & 0 & 0 & 0 & 0 & + & 0 & $-$ & + & $-$ & + & + & + & 0 & + & 0 & 0 & 0 \\
\text{OY} & + & $-$ & + & 0 & 0 & 0 & 0 & 0 & 0 & 0 & 0 & + & 0 & $-$ & + & $-$ & + & + & + & 0 & + & 0 & 0 & 0 \\
\text{P} & $-$ & + & $-$ & 0 & + & 0 & 0 & 0 & + & 0 & 0 & $-$ & 0 & 0 & 0 & 0 & 0 & 0 & 0 & 0 & 0 & 0 & 0 & 0 \\
\text{R} & $-$ & + & + & + & 0 & 0 & 0 & 0 & 0 & + & 0 & + & 0 & 0 & 0 & 0 & 0 & 0 & 0 & 0 & 0 & 0 & 0 & 0 \\
\text{S} & $-$ & + & $-$ & 0 & 0 & 0 & + & 0 & 0 & + & 0 & $-$ & 0 & 0 & 0 & 0 & 0 & 0 & 0 & 0 & 0 & 0 & 0 & 0 \\
\text{SH} & $-$ & + & $-$ & 0 & 0 & 0 & + & 0 & 0 & + & 0 & $-$ & 0 & 0 & 0 & 0 & 0 & 0 & 0 & 0 & 0 & 0 & 0 & 0 \\
\text{T} & $-$ & + & $-$ & 0 & + & 0 & 0 & 0 & 0 & + & 0 & $-$ & 0 & 0 & 0 & 0 & 0 & 0 & 0 & 0 & 0 & 0 & 0 & 0 \\
\text{TH} & $-$ & + & $-$ & 0 & 0 & 0 & + & 0 & 0 & + & 0 & $-$ & 0 & 0 & 0 & 0 & 0 & 0 & 0 & 0 & 0 & 0 & 0 & 0 \\
\text{UH} & + & $-$ & + & 0 & 0 & 0 & 0 & 0 & 0 & 0 & 0 & + & 0 & + & $-$ & $-$ & + & + & $-$ & 0 & 0 & 0 & 0 & 0 \\
\text{UW} & + & $-$ & + & 0 & 0 & 0 & 0 & 0 & 0 & 0 & 0 & + & 0 & + & $-$ & $-$ & + & + & + & 0 & 0 & 0 & 0 & 0 \\
\text{V} & $-$ & + & $-$ & 0 & 0 & 0 & + & 0 & + & 0 & 0 & + & 0 & 0 & 0 & 0 & 0 & 0 & 0 & 0 & 0 & 0 & 0 & 0 \\
\text{W} & $-$ & + & + & + & 0 & 0 & 0 & 0 & + & 0 & + & + & 0 & 0 & 0 & 0 & 0 & 0 & 0 & 0 & 0 & 0 & 0 & 0 \\
\text{Y} & $-$ & + & + & + & 0 & 0 & 0 & 0 & 0 & 0 & + & + & 0 & 0 & 0 & 0 & 0 & 0 & 0 & 0 & 0 & 0 & 0 & 0 \\
\text{Z} & $-$ & + & $-$ & 0 & 0 & 0 & + & 0 & 0 & + & 0 & + & 0 & 0 & 0 & 0 & 0 & 0 & 0 & 0 & 0 & 0 & 0 & 0 \\
\text{ZH} & $-$ & + & $-$ & 0 & 0 & 0 & + & 0 & 0 & + & 0 & + & 0 & 0 & 0 & 0 & 0 & 0 & 0 & 0 & 0 & 0 & 0 & 0 \\
\hline
\end{tabular}
\end{sidewaystable*}

\newpage

\label{sec:appendix}
\clearpage
\appendix

\end{document}